\documentclass[12pt]{article}

\usepackage{setspace}
\usepackage{design_ASC}
\usepackage{amsmath}
\usepackage{caption} % For caption spacing
\usepackage{graphicx}
\usepackage{refstyle}
\usepackage{url}

\usepackage{breakurl}
\usepackage[noadjust]{cite}
\usepackage{geometry}
\usepackage{pdflscape}
\usepackage{authblk}
\usepackage{multirow}
\usepackage{subfigure}% Support for small, `sub' figures and tables

\title{Analysis of the Effectiveness of Face-Coverings on the Death Ratio of COVID-19 Using Machine Learning} %% Assignment Title

\author[1]{Ali Lafzi*}
\author[2]{Miad Boodaghi}
\author[2]{Siavash Zamani}
\author[3]{Niyousha Mohammadshafie}
\author[2]{Veeraraghava Raju Hasti}

\affil[1]{\small Department of Agricultural and Biological Engineering, Purdue University, Indiana 47907, USA}
\affil[2]{\small School of Mechanical Engineering, Purdue University, Indiana 47907, USA}
\affil[3]{\small Department of Civil and Environmental Engineering, University of Pittsburgh, Pennsylvania 15261, USA}

\date{} %% Change "\today" by another date manually
\begin{document}
\setlength{\droptitle}{-5em}    
%% %%%%%%%%%%%%%%%%%%%%%%%%%
\maketitle

% --------------------------
% Start here
% --------------------------
%TC:ignore
\begin{abstract}
\noindent
%(word count = 3644)
The recent outbreak of the COVID-19 led  to the death of millions of people worldwide. To stave off the spread of the virus, the authorities in the US employed different strategies, including the mask mandate order issued by the states' governors. In the current work, we defined a parameter called the average death ratio as the monthly average of the number of daily deaths to the monthly average number of daily cases. We utilized survey data to quantify people's abidance by the mask mandate order. Additionally, we implicitly addressed the extent to which people abide by the mask mandate order that may depend on some parameters like population, income, and education level. Using different machine learning classification algorithms, we investigated how the decrease or increase in death ratio for the counties in the US West Coast correlates with the input parameters. The results showed that for most counties there, the mask mandate order decreased the death ratio reflecting the effectiveness of this preventive measure on the West Coast. Additionally, the changes in the death ratio demonstrated a noticeable correlation with the socio-economic condition of each county. Moreover, the results showed a promising classification accuracy score as high as around 90\%. 
\end{abstract}
%TC:endignore
\section{Introduction}
The recent COVID-19 pandemic has affected millions of people worldwide and led to the tragic death of many innocent lives. The lack of a certain treatment at the beginning of the pandemic traumatized the populace. The only solutions were limited to preventive actions such as wearing face coverings, maintaining social distancing, washing hands, and self-quarantine. Owing to the high transmission rate, only in the US, the number of new daily cases increased from 6 to 22562 during March 2020 according to CDC (Center for Disease Control and Prevention) \cite{CDCOnline}. There is still extensive ongoing research about the possible factors being effective in the pace of this spread; as of now, scientists have declared that meteorological factors such as temperature, wind speed, precipitation, and humidity are some of the critical environmental parameters in this regard \cite{shakil2020covid
}. However, controlling environmental factors involved in the spread of COVID-19 are very challenging and sometimes impossible. As a result, state officials began to impose legislative guidelines, including mandatory use of masks and closure of businesses such as bars and restaurants. Shutting down different businesses has been sporadic due to its adverse economic impact, but obligatory face coverings order is still in effect across the US. In this respect, the effectiveness of facial masks gains further importance and requires scientific studies.
\newline
\\
Presenting a model that can measure the effectiveness of the mask mandate orders can pave the way for governments to take decisive actions during pandemics. The experimental data in tandem with mathematical modelings can be utilized to study the effects of facial coverings on the spread of viral infections. Many previous publications have tried to address the effectiveness of nonpharmaceutical interventions (NPIs) during pandemics, particularly for the spread of influenza  \cite{aiello2010research,saunders2017effectiveness}. Deterministic models have been widely used to study the effects of facial masks on the reproduction number $R_0$. Indeed, the face mask is taken into account by its role in reducing the transmission per contact \cite{brienen2010effect}. The results of the deterministic model indicated that public use of face masks delays the influenza pandemic. On the other hand, some studies suggest that the use of a face mask does not substantially affect  influenza transmission and there is little evidence in favor of the effectiveness of facial masks \cite{xiao2020nonpharmaceutical,cowling2010face}. As for the COVID-19, the efficacy of the facial mask in impeding the infectivity of the SARS-CoV-2 remains unclear. Considering the effects of mask in reproduction number $R_0$, Li et al. \cite{li2020mask} claimed that wearing face masks alongside the social distancing can flatten the epidemic curve. Other studies also pinpointed that public use of a facial mask may reduce the spread of COVID-19 \cite{cheng2020role}. Despite these findings, the efficacy of face masks remains controversial.\newline 
\\
The cardinal point that has not garnered enough attention is the relationship between the degree of exposure to the virus and its mortality rate. Some researchers presented the idea that the severity of the symptoms correlates with the extent of exposure to justify the high death rate in healthcare workers \cite{loadOnline}. Unfortunately, there is no universal trend that can predict the relationship between the dose of the virus and the severity of the resulting symptoms. A study performed on the relationship between influenza and rhinovirus viral load and the severity in the upper respiratory tract infections reported a different behavior for those viruses \cite{granados2017influenza}. In fact, the results indicated that for influenza A and the rhinovirus, viral loads were not associated with hospitalization/ICU. On the other hand, for influenza B, viral load was higher in hospitalized/ICU patients. Furthermore, for Respiratory syncytial virus (RSV),  the viral load seems to correlate with the severity of symptoms as many studies in the literature suggest that a correlation exists \cite{martin2008clinical,houben2010disease,devincenzo2005respiratory}. The same controversy holds for the COVID-19. Recently, some studies have tried to investigate the severity of COVID-19 with its load, where they found that the load tightly correlates with the severity \cite{liu2020correlation,fajnzylber2020sars}. However, another study suggests that no such a correlation exists \cite{he2020temporal}.
\newline
\\
To unveil whether COVID-19 viral load is related to disease severity requires an in-depth study, which involves infecting volunteers with controlled doses of virus and monitoring their symptoms. However, experimental challenges, in addition to the ethicality of these experiments, make this type of research very challenging \cite{loadOnline}. Although studies have not been convergent in whether nose \cite{hou2020sars} or mouth \cite{huang2020integrated} is the primary site for COVID-19 infection, they underscored the importance of wearing a facial mask as a barrier to the virus spread. Additionally, although the protection level of different types of mask is different, wearing any mask, even a cloth mask, is better than wearing nothing at all, which can play a role in protection from the exposure to COVID-19 \cite{goh2020face,sharma2020efficacy}. As mentioned, conducting experimental studies to reveal the relationship between the extent of exposure and severity of COVID-19 is very challenging. One way to circumvent theses challenges is to conduct an indirect study by introducing a model that can capture changes in the mortality rate due to the wearing a facial mask. Indeed, if the ratio of the number of death to the number of cases decreases, this can support the hypothesis that there is a correlation between the viral load and the severity of symptoms. Thus, studying the effects of Mask Mandate order on the mortality rate gains extra importance.
\newline
\\
A Machine Learning (ML) analysis can be instrumental in shedding light on the possible correlation between the public use of masks and changes in the mortality rate. The success of implementing ML and Artificial Intelligence (AI) techniques in the previous pandemics has convinced researchers to use them as precious tools in fighting against the current outbreak \cite{lalmuanawma2020applications}. ML and AI can be used for prediction and forecasting in different regions so that the corresponding health officials can take necessary actions in advance \cite{lalmuanawma2020applications}. In addition, this technology is capable of enhancing the prediction accuracy for screening both infectious and non-infectious diseases \cite{agrebi2020use}. Six ML methods have been carried out to predict $1$, $3$, and $6$ days ahead the total number of confirmed COVID-19 cases with errors in the ranges of 0.87\%–3.51\%, 1.02\%–5.63\%, and 0.95\%–6.90\%, respectively, in 10 Brazilian states \cite{ribeiro2020short}. Moreover, an ML method like XGBoost model was capable of identifying $3$ important biomarkers from $485$ blood samples in Wuhan, China as the key mortality parameters \cite{yan2020interpretable}. ML algorithms also have been used to capture the correlation between the weather data and COVID-19 mortality and transmission rates   \cite{malki2020association, shrivastav2020gradient}. Additionally, ML has been utilized to study the effects of mask mandate (MM) order on the number of daily cases, where no significant statistical difference was observed in the number of daily cases in the state-wise analysis \cite{maloney2020mask}. These studies confirm the strength of ML as a great tool to investigate the effects of MM order on mortality rates of COVID-19.   
\newline
\\
Another important factor regarding the effectiveness of MM order is society's adherence to the regulations. One study that tried to quantify public compliance with COVID-19 public health recommendations found notable regional differences in intent to follow health guidelines \cite{lennon2020public}. In addition, some studies noticed a correlation between the level of education and intent to voluntarily adhere to social distancing guidelines \cite{lennon2020public,sathianathan2020knowledge}. However, not only the level of education but also the level of income and race can play a role in the adherence to the regulations \cite{weiss2020disparities}. Based on these findings, it's important to take into account the features that might be correlated with people's compliance with the MM order. Additionally, we will use data based on the survey provided by the New York times available on Github, which quantifies people's adherence to the MM order \cite{nymaskOnline}. As a result, in this study, we will include factors that might play a role in people's adherence to the MM order as our input features.
\newline 
\\
In the proposed work, utilizing different ML classification algorithms, we aim to unveil how the change in the mortality rate correlates with certain features. The features will be chosen in a way that they can reflect abidance by MM order in different counties. We will use the data provided by CDC to find the average monthly number of COVID-19 cases. Additionally, the exact dates of the executive orders signed by the state officials are available for each state - California: June 18th 2020, Oregon: June 19th 2020, Washington: June 26th 2020. To have appropriate unbiased data, similar to what Maloney et al. \cite{maloney2020mask} has done in his study of the effect of mask mandate, we will be using the data for one month after and before the executive orders for each preventive measure for the three states on US West Coast. With this data selection method, we limit the geographical region of the study to ensure that changes in the cases are highly attributed to the public use of masks rather than other factors such as environmental changes. \newline \\
The rest of the paper is organized as follows. First, we will represent how our data was collected and arranged. Then we will explain the ML methods we have used for our prediction. Finally, we will describe and compare the results obtained from different ML methods.

\section{Methodology}
In this section, we will explain the collected data and the  ML algorithms used for the training and prediction.

\subsection{Data} \label{data_collection}
\begin{figure}[h]
    \centering
    \resizebox*{7cm}{!}{\includegraphics{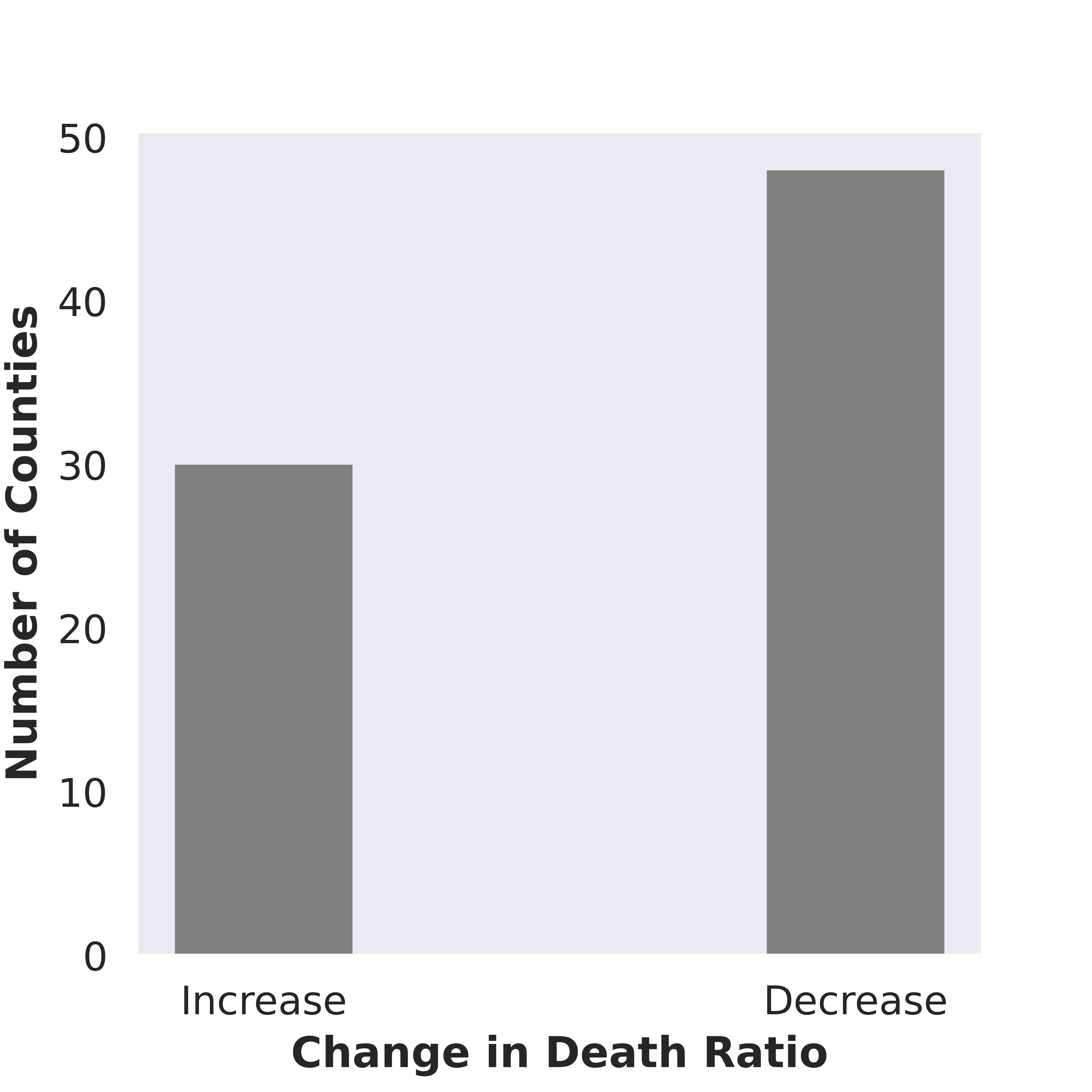}}
    \caption{Histogram of change in death ratio for the three states}
    \label{classes_histogram}
\end{figure}

We defined the parameter of interest as the ratio of the monthly average number of deaths to the monthly average number of cases, referred to as the death ratio, which can be interpreted as a measure of the severity of the disease.
The effective date of the executive orders by the governors, requiring mask mandate at all the counties in the three West Coast states of California, Oregon, and Washington, has been identified, which is publicly available \cite{orderdate}. We used the average death ratio one month before and after the order to study the mortality rate. The rationale behind this selection is to minimize the effects of other factors that might play a role in changing the COVID-19 data. The raw dataset for the daily cases and deaths for all the US counties over time is extracted from the USAFACTS website \cite{USAfacts}, where county-level data is confirmed by the state and local agencies directly. After obtaining the daily values of death and case numbers for a month before and after the MM order, we divided the monthly average number of deaths by the monthly average number of cases for each county. Then we found the difference between the death ratio for one month before and after the MM order. Finally, we categorized the variation based on its sign to quantify whether the death ratio increases, decreases, or no change occurs. Out of the 130 samples, 47, 30, and 53 of them belong to the ``decrease'', ``increase'', and ``no change'' classes, respectively. We dropped the ``no change'' data as they all correspond to small counties, where there were zero reported COVID-19 cases and deaths, leaving 77 counties in total. Consequently, the two categories of increase and decrease in the death ratio remain for the prediction task. Figure \ref{classes_histogram} illustrates the number of samples in each category, which expresses that the available data for classification is not biased. %The single row corresponding to the county where the death rate was negative was removed.
\\
\newline
It is a hard task to directly determine the exact percentage of the population that follows the MM order and uses face coverings. As a result, it is necessary to come up with features that can indirectly capture how likely is an individual to follow the recommended practice. For bridging this gap, four main features are chosen as primary indicators, which are listed below:
\begin{enumerate}
    \item County population
    \item Median household income
    \item Education level
    \item Mask usage based on New York Times survey
\end{enumerate}
Population in each county is obtained from the most recent surveys for the year 2019. The income level is the median household income in US dollars and the education level is the percentage of people who have completed high school in each county in the years 2015-2019. The raw data for these features is  obtained from the US Census website \cite{USCensus}. The US Census measures the median income as the regular income received, excluding other payments like tax, etc \cite{USCensus_income}. Furthermore, we used survey data provided by the New York Times that quantifies the mask usage from 7/2/2020 to 7/14/2020 \cite{nymaskOnline}. Since the survey timeline lies within the month after the MM order for all three studied states, it is valid to use its data for our purpose. Finally, we will try to establish an AI-based relationship between the features and the sign of the change in the death ratios of the Pacific Coast states at the county level using nine different classification algorithms, provided in section \ref{ML_methods}. 

%The step by step process is presented below:
%\begin{itemize}
%    \item Find the Mask Mandate (MM) order's effective date for the states of interest.
 %   \item Find the COVID-19 data daily cases and deaths data for a month before and after the MM order.
  %  \item Find the data for the states of interest.
   % \item Read the data using Pandas.
%    \item Categorize the data based on the sign of the variation in death ratio.
%\end{itemize}
\subsection{Methods} \label{ML_methods}
In this study, we have developed machine learning models to correlate the specified features mentioned in section \ref{data_collection} to shed light on the relationship between adherence to mask mandate and mortality rate.

Classic ML methods of Logistic Regression \cite{ayyadevara2018pro} and Naive Bayes classifier \cite{richert2013building} are used. In addition, ensemble learning-based models, Random Forest and Extra Trees, are also analyzed \cite{steinki2015introduction}. Moreover, the extreme boosting method, XGBoost is explored \cite{liang2020predicting}. Other methods such as Support Vector Machine, K-Nearest Neighbors \cite{gad2020comparative}, Decision Trees \cite{priyanka2020decision}, and Neural Network \cite{deng2018deep} are additionally used for prediction of effect of Mask Mandate on mortality rate.

 It should be noted that for carrying out the analysis, the data is split into training and test sets, with a test size of 20\% \cite{khuzani2021covid}. A k-fold cross-validation scheme with five folds has been used to evaluate the performance of each method on the validation set and tune its hyper-parameters with the classification accuracy as the metric accordingly. The hyper-parameter tuning is done using either grid search or random search for all the methods. A statistical summary of the final dataset for binary classification is outlined in table \ref{dataset_statistics}, which indicates a significant difference between the orders of magnitudes of the features. Therefore, min-max and max-abs scaling have been used to transform the input features and output, respectively, before passing the data to the ML algorithms for training. It should be noted that the data used in this article were access through publicly available sources as listed, and we confirm that all methods were performed in accordance with the relevant guidelines and regulations.

\begin{table}[h]
\caption{Statistical summary of the final dataset before scaling. Columns are P:population, MI:median income, EL:education level. Mask usage - N:never, R:rarely, S:sometimes, F:frequently, A:always. DR:change in death ratio between one month before and after the corresponding MM order date}
\begin{tabular}{|l|l|l|l|l|l|l|l|l|l|l|}
\hline
\multirow{2}{*}{} & \multicolumn{1}{c|}{\multirow{2}{*}{\textbf{P}}} & \multicolumn{1}{c|}{\multirow{2}{*}{\textbf{MI}}} & \multicolumn{1}{c|}{\multirow{2}{*}{\textbf{EL}}}  & \multicolumn{5}{c|}{\textbf{Mask Usage}}    & \multicolumn{1}{c|}{\multirow{2}{*}{\textbf{DR(\%)}}} \\ \cline{5-9}
                                    & \multicolumn{1}{c|}{}                    & \multicolumn{1}{c|}{}                     & \multicolumn{1}{c|}{}                     & \textbf{N}     & \textbf{R}    & \textbf{S}     & \textbf{F}    & \textbf{A}    & \multicolumn{1}{c|}{}                        \\ \hline
\textbf{Count}             & 77                                      & 77                                       & 77                                                                                & 77    & 77   & 77    & 77   & 77   & 77                                           \\ \hline
\textbf{Mean}              & 630413.5                                & 66494.23                                 & 0.85                                                                            & 0.03  & 0.03 & 0.06  & 0.17 & 0.71 & -0.47                                        \\ \hline
\textbf{Std}               & 1297275                                 & 18484.92                                 & 0.07                                                                              & 0.02  & 0.03 & 0.03  & 0.05 & 0.09 & 2.83                                         \\ \hline
\textbf{Min}               & 7208                                    & 43313                                    & 0.67                                                                                  & 0.001 & 0    & 0.004 & 0.07 & 0.31 & -12.9                                        \\ \hline
\textbf{25\%}              & 86085                                   & 53105                                    & 0.81                                                                                  & 0.02  & 0.01 & 0.04  & 0.14 & 0.67 & -1.4                                         \\ \hline
\textbf{50\%}              & 219186                                  & 62077                                    & 0.88                                                                                  & 0.02  & 0.02 & 0.06  & 0.16 & 0.72 & -0.44                                        \\ \hline
\textbf{75\%}              & 601592                                  & 74624                                    & 0.91                                                                                  & 0.04  & 0.04 & 0.08  & 0.2  & 0.77 & 0.77                                         \\ \hline
\textbf{Max}               & 10039110                                & 124055                                   & 0.96                                                                                  & 0.11  & 0.21 & 0.21  & 0.3  & 0.87 & 7.69                                         \\ \hline
\end{tabular}
\label{dataset_statistics}
\end{table}

\section{Results and Discussions}
\begin{figure}[!htb]
\centering
    \resizebox*{12cm}{!}{\includegraphics{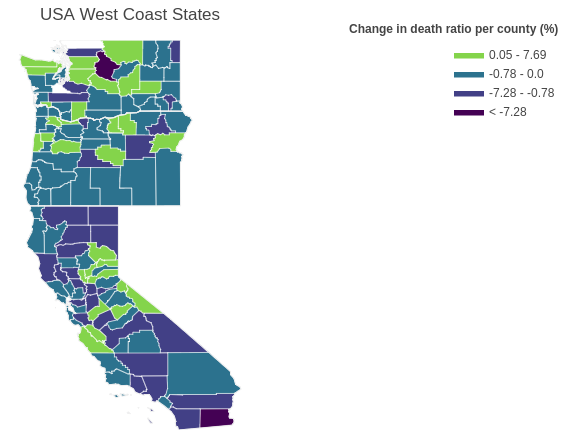}}
    \caption{Change in death ratio in US West Coast states counties}
    \label{DR_map}
\end{figure}

The change in death ratio from one month before to one month after the date of mandating face-covering in the three states is visualized for each county in fig. \ref{DR_map}. Two clusters of increase in death ratio can be seen, one near northern Washington and one near central California. Our first intuition was that by increasing the population, the chance of viral spread would increase. Therefore, we expected to see a positive change in the death ratio for more populated counties. However, as it can be seen from the map, there is inherent randomness that defies our initial intuition about the spread mechanism. Further, it is shown that more counties experienced a decrease in death ratio one month after the usage of face-covering was mandated by each state, as shown in fig. \ref{classes_histogram}. Therefore, usage of face-covering is chosen as the main factor affecting the decrease of the change in death ratio. As explained previously, to quantify adherence to the mask mandate, other auxiliary features are chosen, namely, median income, and education level for each county.

\begin{figure}[h]
    \centering
    \resizebox*{7cm}{!}{\includegraphics{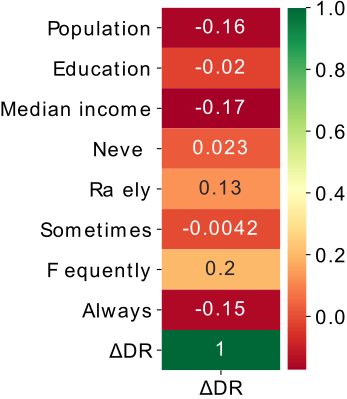}}
    \caption{Correlations between the features and the output}
    \label{feature_corr}
\end{figure}

The combined effect of features is analyzed on the death ratio. Then the performance of each algorithm is evaluated for test and train sets. The effect of each feature on the change of death ratio is visualized by the correlation heatmap provided in the fig. \ref{feature_corr}. Here, we have not presented the cross-correlations between features for a simpler visualization. Each row of the complete correlation matrix is an appropriate indicator of how correlated that feature is with the change in death ratio, which is what fig. \ref{feature_corr} illustrates. A more negative value implies that the increase of that specific feature is positively correlated by a decrease in the change of death ratio.
For instance,  an increase in population, median income, and education level would result in a decrease in the change of death ratio. %On the other hand, the positive correlation for republican votes leads to a more significant increase in the death ratio.
An interesting observation is the disordered correlation pattern for mask usage. As one expects, increasing the number of never and rarely mask users is positively correlated with a change in the death ratio. However, the data associated with frequently mask users have resulted in a positive correlation value. Such erratic correlation behavior necessitates the inclusion of other features in the analysis.

%The results of analysis for each different algorithm to predict whether a county experiences increase or decrease in the death ratios is provided in table \ref{results_table}.
%For instance, a high accuracy score means that there is a strong correlation between the all the features and the death ratio, and therefore, the obtained parameters could be used to inspect how the features affect the death ratio.

\begin{figure}[h] 
    \centering
    \subfigure[]{
    \includegraphics[height=5cm]{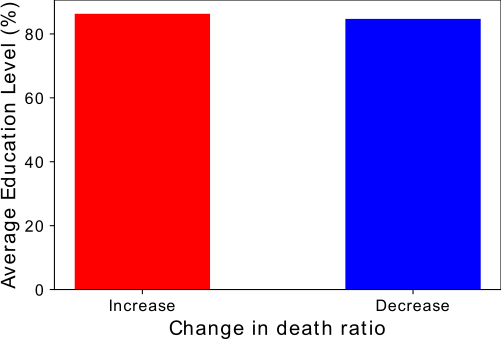}
    \label{bar_education}}
    \subfigure[]{
    \includegraphics[height=5cm]{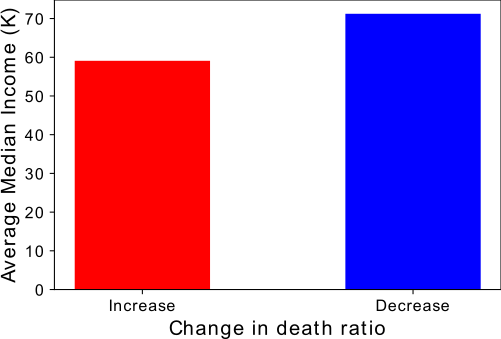}
    \label{bar_income}}
    \subfigure[]{
    \includegraphics[height=5cm]{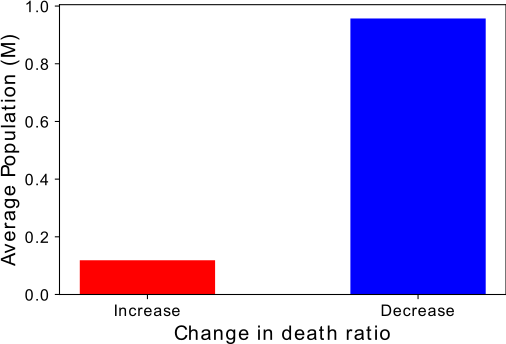}
    \label{bar_population}}
    \caption{Visualization of the combined data for California, Oregon and Washington. Change in death ratio and average of (a) education level (b) median income and (c) population.}
    \label{BarGraph}
\end{figure}

As a preliminary analysis, the relationship between the average values of the three auxiliary features and the change in death ratio has been visualized for each category separately in fig. \ref{BarGraph}. Figure \ref{bar_education} expresses that the average percentages of people who have completed high school education in both categories of counties that have experienced an increase or decrease in their death ratios are almost the same. This could indicate why the correlation between this feature and output is very close to zero, as represented in fig. \ref{feature_corr}. Further, a noticeable correlation is observed between average median income and the change of death ratio, presented in fig. \ref{bar_income}. On average, the communities with less median income experienced a positive change in death ratio, meaning more mortality rate, which is in agreement with what is reported in \cite{weiss2020disparities}. However, the strongest correlation is observed by considering county population, shown in fig. \ref{bar_population}. The counties with fewer residents were affected more adversely by the pandemic compared to high-population counties. The counterintuitive relation between population and change in death ratio further corroborates the necessity of inclusion of the two other supplementary features.
%As a preliminary analysis, political inclination, based on the 2020 presidential election, is chosen as the focal criterion to categorize the data for changes of death ratio for all three states, as presented in Fig. (\ref{BarGraph}). Fig. \ref{bar_political} shows that in general, communities that voted republican in the presidential election of 2020 were affected worse compared to democratic counties. 

\begin{figure}[!htb]
    \centering
    \resizebox*{13.5cm}{!}{\includegraphics{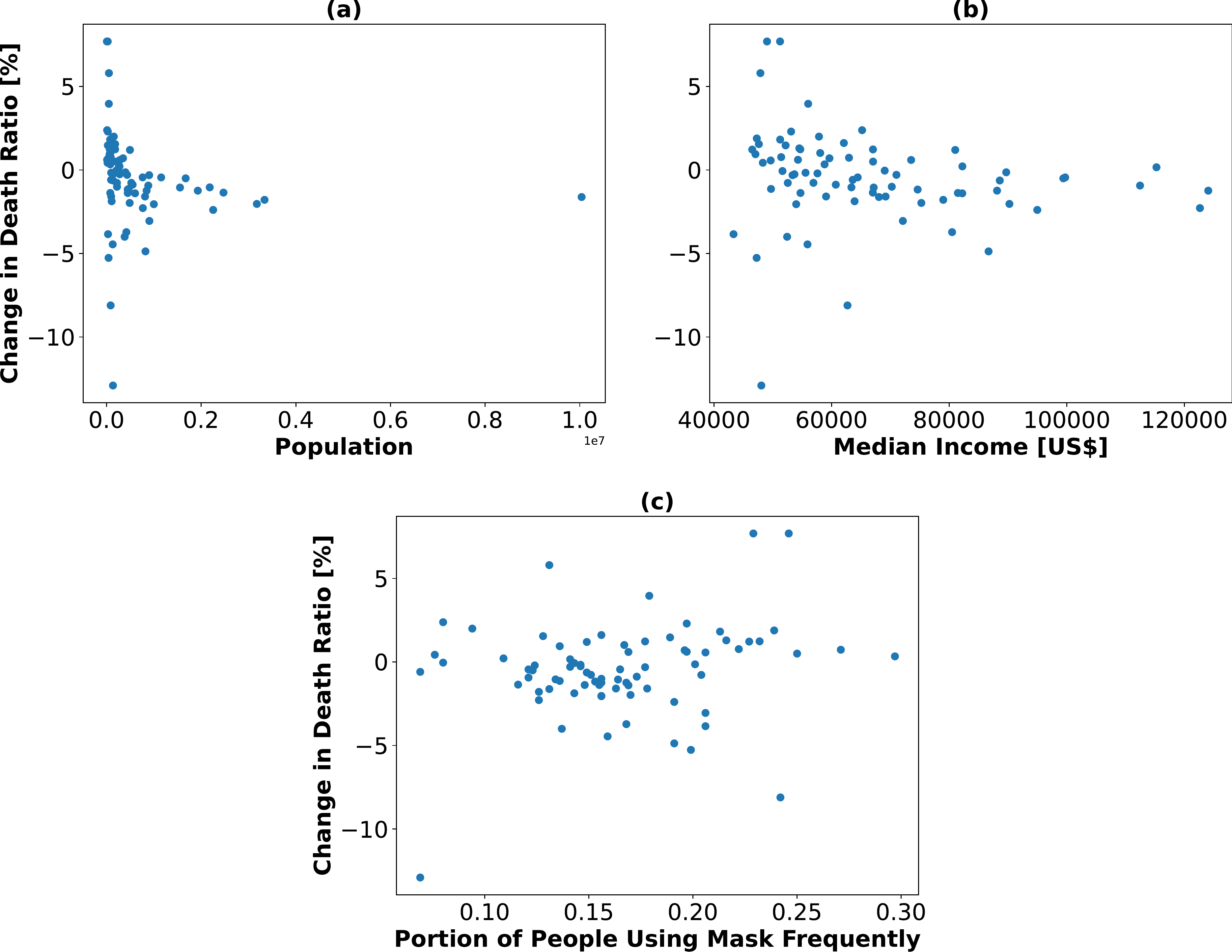}}
    \caption{Scatter plot of the percent change in the death ratio as a function of a) population b) median income c) percent people frequently using masks.}
    \label{scatter_all}
\end{figure}

To have an initial assessment of the variation of percent change in the death ratio, we plotted the percent death ratio as functions of population, median income, and percent of the population that frequently uses mask, which has a relatively high correlation coefficient according to fig. \ref{feature_corr}. Figure \ref{scatter_all} a-c shows no detectable pattern between parameters of interest and death ratio. As a result, it is not possible to predict the value of change in the death ratio using regression. On the other hand, as we will show, converting changes to categories of increase and decrease would pave the way for capturing the status of the change.

A summary of the overall death ratios in the months before and after the mask mandate order for the three states is presented in table \ref{death_rates}. It can be observed that death ratio significantly decreases in California and Washington but slightly increases in Oregon. This result suggests an intrinsically complex pattern between the death ratio as the output and the selected inputs. Table \ref{deaths_cases_change} shows the changes in the average number of deaths and cases between the months before and after the MM order within the entire states. It can be seen that while the average number of deaths has decreased in Washington and increased in Oregon and California, the average number of cases has increased in all of them. This implies that the observed decrease in death ratios, as reported in table \ref{death_rates}, can be because of the effect of face coverings in reducing the severity of COVID-19 infection.

\begin{table}[!htb]
\centering
\caption{Total death ratios in the month before and after the corresponding date of the mandatory face coverings executive order in each state}
\begin{tabular}{llll}
\hline
\textbf{State}                 & \textbf{1 month before MM order} & \textbf{1 month after MM order} & \textbf{Change} (\%) \\
\hline
California            & 63.13           & 32.67 & -48\\
Washington            & 28.16         & 21.15             & -25\\
Oregon & 38.03         & 39.14              & +3  \\
\hline
\end{tabular}
\label{death_rates}
\end{table}

\begin{table}[!htb]
\centering
\caption{Changes in the average cases and deaths between one month before and after the MM order across the entire states}
\begin{tabular}{llll}
\hline
\textbf{State}                 & \textbf{Changes in the average cases} & \textbf{Changes in the average deaths}  \\
\hline
California            & 4210.74          & 9.76 \\
Washington            & 379.56         & -0.82           \\
Oregon & 172.85         & 0.75               \\
\hline
\end{tabular}
\label{deaths_cases_change}
\end{table}

\begin{figure}[!htb] 
    \centering
    \subfigure[]{
    \includegraphics[height=4cm]{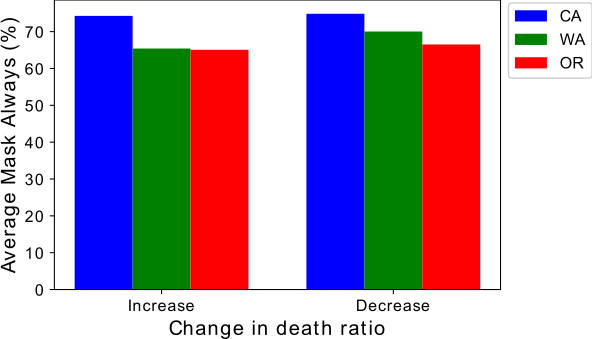}
    \label{mask_always}}
    \subfigure[]{
    \includegraphics[height=4cm]{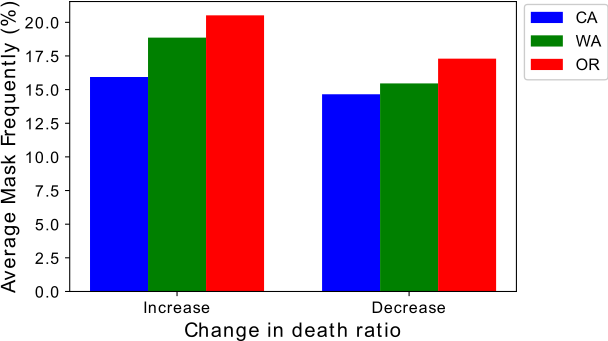}
    \label{mask_frequently}}
    \subfigure[]{
    \includegraphics[height=4cm]{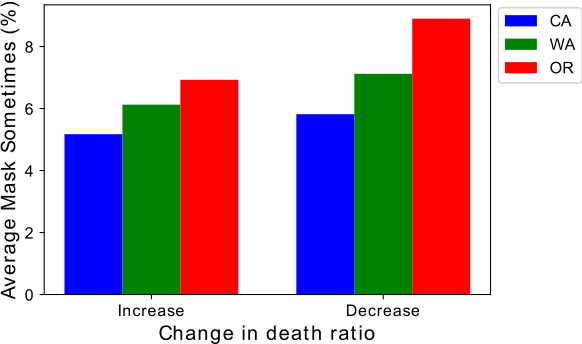}
    \label{mask_sometimes}}
    \subfigure[]{
    \includegraphics[height=4cm]{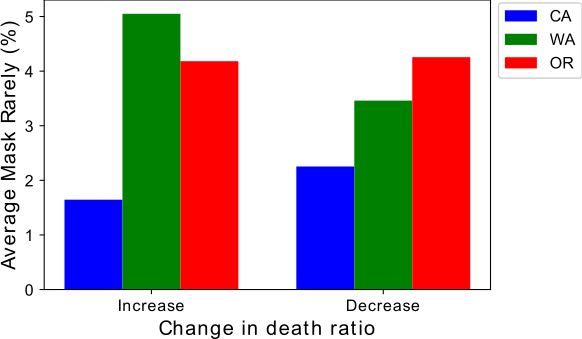}
    \label{mask_rarely}}
    \subfigure[]{
    \includegraphics[height=4cm]{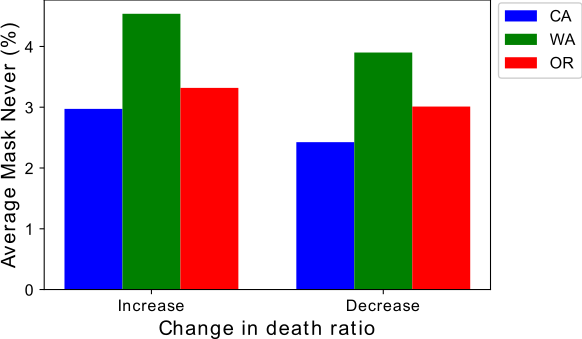}
    \label{mask_never}}
    \caption{Average percentage of people who (a) always, (b) frequently, (c) sometimes, (d) rarely, and (e) never use mask across all the counties experiencing an increase and decrease in their death ratios}
    \label{mask_rates}
\end{figure}

Furthermore, to get some insight about the observed pattern in table \ref{death_rates}, the average percentage of people who use masks with different frequencies across all the counties experiencing both increase and decrease in their death ratios is illustrated in fig. \ref{mask_rates}. As expected, the average percentage of people who always wear a mask (fig. \ref{mask_always}) is slightly higher for the decreasing category, but in both categories, the values are the smallest for Oregon. Moreover, the average percentage of people who never use a mask (fig. \ref{mask_never}) is lower for the decreasing category in all 3 states, which is also intuitively sensible. Although there are no prominent and consistent patterns for the remaining mask usage frequencies, these observations could implicitly and partially describe why there is a small increase in the overall death ratio in Oregon. However, since the mask usage data is from an NYT survey over a limited period (12 days), the observations in fig. \ref{mask_rates} cannot explain the entire underlying phenomenon. According to a recent study, several factors are attributing to the possibility of a person following or not following the health guidelines set by the state officials \cite{weiss2020disparities}. Therefore, three features among these parameters plus the aforementioned mask usage as the fourth feature have been used to conduct the current study.

%\begin{figure}[h]
%    \centering
%    \resizebox*{12cm}{!}{\includegraphics{Graphs/mask_always_barplot.eps}}
%    \caption{Average percentage of people who always use mask across all the counties experiencing an increase and decrease in their death ratios}
%    \label{mask_always}
%\end{figure}

All implemented algorithms in this study are capable of providing us with high classification accuracy, i.e,  of predicting whether a county has experienced a decrease in its death ratio after the MM order or an increase. As provided in table \ref{accuracy_results}, it can be seen that, in general, most of the algorithms have relatively high accuracy scores for the test set. Despite the lack of sufficient training data set, Naive Bayes has an accuracy of 94\%, and Random Forest, XGBoost, and Decision Tree have an accuracy of 88\%. The selected hyper-parameters for XGBoost, Decision Tree, and Random Forest classifiers are shown in table \ref{hyperparameters}. The random search method has been done to tune these hyper-parameters for XGBoost, and grid search is used for Random Forest and Decicion Tree. Naive Bayes does not have any important hyper-parameter because of which, it has the capability of being generalized well. Besides, Random Forest, as a bagging method, and XGBoost, as a boosting method, have the popularity of rarely over-fitting the data. Moreover, the final hyper-parameters after tuning for the rest of the implemented algorithms are given in table \ref{remaining_hyperparameters}. Except for KNN where the elbow method is used to find the optimum number of nearest neighbors, the grid search method is applied for hyper-parameter tuning of the other methods in this table. Additionally, table \ref{accuracy_results} includes the 95\% confidence interval (denoted by CI (\%)) for all algorithms. The interval is calculated based on the following:
\begin{equation}\label{confidence_interval}
CI=z\sqrt\frac{score(1-score)}{n_{test}},
\end{equation}
Where $z$ is the number of standard deviations from the Gaussian distribution and equals to 1.96 for 95\% CI, score is the classification accuracy of the algorithm, and $n_{test}$ is the number of test points in our dataset. As expected, we see this interval becomes narrower as the test set accuracy increases. By taking a closer look at the accuracy scores on the train set and comparing them with those of the test set, we note that none of the implemented methods overfits the data.%These reasons could be why these three algorithms have outperformed the others.
\begin{table}[!htb] 
\centering
\caption{Performance metrics for all the studied algorithms.}
\begin{tabular}{|l|c|c|c|}
\hline
\textbf{Algorithm}     & \textbf{Test(\%)} &\textbf{Train(\%)} &\textbf{CI(\%)} \\ \hline
Support Vector Machine                     & 69 & 74 & 23                   \\ \hline
Extra Trees                                & 75 & 93& 21                  \\ \hline
KNN                                       & 81 &75 &19                  \\ \hline
Logistic Regression                        & 81  & 79 & 19                 \\ \hline
Neural Net                                 & 81 & 80 & 19                  \\ \hline
Decision Tree                             & 88    & 93& 16               \\ \hline
Random Forest                              & 88 & 85& 16                  \\ \hline
XGBoost                                    & 88 & 95 & 16                  \\ \hline
Naive Bayes                                 & 94  & 70& 12                  \\ \hline
\end{tabular}
\label{accuracy_results}
\end{table}

\begin{figure}[!htb]
    \centering
    \resizebox*{12cm}{!}{\includegraphics{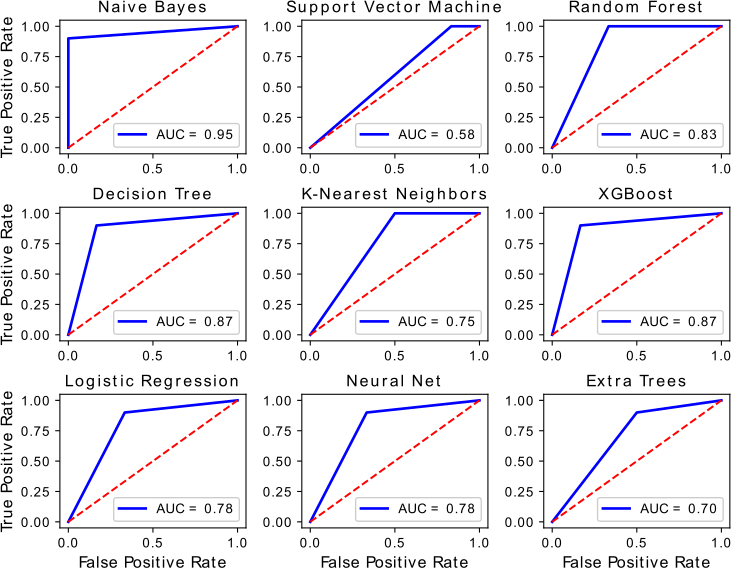}}
    \caption{ROC plots of all the algorithms}
    \label{ROC}
\end{figure}

Additionally, to further compare the predictive power of the algorithms, the receiver operating characteristic ROC curve is shown in fig. \ref{ROC}. The ROC curve shows the performance of a classification model at all classification thresholds. The diagonal red dashed lines demonstrate the no-discrimination line, which corresponds to the values of a random guess. As evident, for all algorithms, the ROC curve is above the line of no-discrimination. The area under ROC Curve (AUC) values demonstrate an aggregate measure of performance across all possible classification thresholds. In other words, the AUC values measure how accurate the model predictions are, and the values close to 1 are desirable. The measured AUC values, as shown in fig. \ref{ROC}, suggest that the Naive Bayes algorithm leads to the best prediction. The AUC values are also in agreement with the testing accuracy, as shown in table \ref{accuracy_results}, where also the Naive Bayes algorithm has the highest testing accuracy.
%\begin{table}[h] 
%\centering
%\caption{Train and test accuracies for all the studied algorithms.}
%\begin{tabular}{|l|l|l|}
%\hline
%\textbf{Algorithm}     & \textbf{Train Accuracy} & \textbf{Test Accuracy} \\ \hline
%Support Vector Machine & 0.82                    & 0.81                   \\ \hline
%Decision Tree          & 1.00                    & 0.81                   \\ \hline
%KNN                    & 0.74                    & 0.69                   \\ \hline
%Logistic Regression    & 0.79                    & 0.75                   \\ \hline
%Neural Net             & 0.78                    & 0.88                   \\ \hline
%Extra Trees            & 0.93                    & 0.81                   \\ \hline
%Naive Bayes            & 0.7                     & 0.94                   \\ \hline
%Random Forest          & 1.00                    & 0.94                   \\ \hline
%XGBoost                & 0.98                    & 0.94                   \\ \hline
%\end{tabular}
%\label{accuracy_results}
%\end{table}

\begin{table}[!htb]
\centering
\caption{Model Parameters for XGBoost, Random Forest, and Decision Tree. Columns of XGBoost - CSbT:column sample by tree ratio, G:gamma, LR:learning rate, MD:maximum depth of each tree, NE:number of estimators, S:subsamples ratio. Columns of Random Forest - MD:maximum depth of each tree, MF:maximum number of features for best split, MSS:minimum number of samples to split an internal node, NE:number of estimators. Columns of Decision Tree - MD:maximum depth of each tree, CR: criterion to measure the quality of a split, MSS:minimum number of samples to split an internal node}
\begin{tabular}{|l|l|l|l|l|l|}
\hline
\multicolumn{6}{|c|}{\textbf{Extreme Gradient Boosting (XGBoost)}}                                                                     \\ \hline
CSbT & G  & LR    & MD   & NE    & S    \\ \hline
0.9605                  & 0.4735 & 0.0975           & 4           & 119                     & 0.6232                     \\ \hline
\end{tabular}
\begin{tabular}{|l|l|l|l|}
\hline
\multicolumn{4}{|c|}{\textbf{Random Forest}}                                                                                 \\ \hline
MD  & MF & MSS & NE \\ \hline
7          & 2   & 2                 &10             \\ \hline
\end{tabular}
\begin{tabular}{|l|l|l|l|}
\hline
\multicolumn{3}{|c|}{\textbf{Decision Tree}}                                                                                 \\ \hline
MD  & CR & MSS  \\ \hline
4             & Gini                 &2             \\ \hline
\end{tabular}
\label{hyperparameters}
\end{table}

\begin{table}[!htb]
\caption{Model Parameters for the remaining methods.}
\centering
\begin{tabular}{cccll}
\hline
\multicolumn{5}{|c|}{\textbf{Neural Network}}                                                                                                   \\ \hline
\multicolumn{1}{|c|}{Activation}     & \multicolumn{1}{c|}{Learning Rate}   & \multicolumn{1}{c|}{Neurons} & \multicolumn{1}{c|}{Layers} & \multicolumn{1}{c|}{Epochs} \\ \hline
\multicolumn{1}{|c|}{ReLU}    & \multicolumn{1}{c|}{0.01} & \multicolumn{1}{c|}{32}      & \multicolumn{1}{c|}{1}  & \multicolumn{1}{c|}{50}     \\ \hline
\multicolumn{1}{l}{}          & \multicolumn{1}{l}{}      & \multicolumn{1}{l}{}         &                         &                             \\ \hline
\multicolumn{5}{|c|}{\textbf{KNN}}                                                                                                               \\ \hline
\multicolumn{5}{|c|}{Nearest Neighbors Number}                                                                                                   \\ \hline
\multicolumn{5}{|c|}{8}                                                                                                                          \\ \hline
\multicolumn{1}{l}{}          & \multicolumn{1}{l}{}      & \multicolumn{1}{l}{}         &                         &                             \\ \hline
\multicolumn{5}{|c|}{\textbf{Logistic Regression}}                                                                                               \\ \hline
\multicolumn{2}{|c|}{C (penalty term)}                                   & \multicolumn{3}{c|}{Regularization Norm}                                             \\ \hline
\multicolumn{2}{|c|}{1000}                                & \multicolumn{3}{c|}{L2}                                                              \\ \hline
\multicolumn{1}{l}{}          & \multicolumn{1}{l}{}      & \multicolumn{1}{l}{}         &                         &                             \\ \hline
\multicolumn{5}{|c|}{\textbf{Extra Trees}}                                                                                                       \\ \hline
\multicolumn{1}{|c|}{Criterion}      & \multicolumn{1}{c|}{Min. Samples Split}  & \multicolumn{1}{c|}{Min. Samples Leaf}     & \multicolumn{2}{c|}{Estimator Count}                 \\ \hline
\multicolumn{1}{|c|}{entropy} & \multicolumn{1}{c|}{2}    & \multicolumn{1}{c|}{2}       & \multicolumn{2}{c|}{200}                              \\ \hline
\multicolumn{1}{l}{}          & \multicolumn{1}{l}{}      & \multicolumn{1}{l}{}         &                         &                             \\ \hline
\multicolumn{5}{|c|}{\textbf{Support Vector Machine}}                                                                                            \\ \hline
\multicolumn{1}{|c|}{C (penalty coefficient)}       & \multicolumn{4}{c|}{Kernel}                                                                                      \\ \hline
\multicolumn{1}{|c|}{1}       & \multicolumn{4}{c|}{Radial Basis Function}                                                                       \\ \hline
\end{tabular}
\label{remaining_hyperparameters}
\end{table}

The trend of high accuracy on test data signifies the existence of a pattern between the chosen features and the change in death ratio in the proposed model. Moreover, against the common belief that highly populated areas might experience harsher effects of COVID-19, on the west coast of the United States, the areas with lower populations endured worse conditions. Additionally, such a modeling approach could be used to optimize the  distribution of services and media coverage for possible future adversities. A possible solution for decreasing the effect of future pandemics such as COVID-19 would be improving media coverage and public knowledge, especially in more vulnerable areas. 

%\clearpage
\section{Conclusion}
In this body of work, we have analyzed the effect of mask covering on the intensity of spread of the COVID-19 virus by considering the death ratio at the county level to be the primary indicator. To bridge the gap between level of adherence to mask mandate, we use four main features as input data: population, income, education level, and the survey results on mask usage from the New York Times. The change in the death ratio is used as the metric to quantify the effectiveness of face-coverings on the COVID-19 spread. After extracting and refining the data-set from reliable sources, we analyzed the information using nine different algorithms. Among all the methods used, Random Forest, XGBoost, Decision Tree, and Naive Bayes had the best performance with a classification accuracy of around 90\%. This high accuracy shows the legibility of chosen features as influential criteria for modeling purposes. The obtained hyper-parameters for these models, along with the selected features, can now be used to predict future conditions of the spread of the virus. 

The results show a connection between adherence to the mask mandate and change in death ratio in most counties studied. The findings of this work emphasize the potential role the immediate legislative action can play in improving the society's well-being during pandemics. It is hoped that the results of this work could further clarify the importance of preventive measures such as MM order and highlight the importance of socioeconomic conditions on the behavior of different communities, which could be complex and counter-intuitive. However, it is important to note that the results we presented here are valid only for a specific geographic location, which in this study was the West Coast of the United States. Any generalization of our findings that can be interpreted as guidelines bears over-arching and worldwide studies.    

%TC:ignore
%\section{Disclaimer}
%None of the authors of this work have had any political affiliation or interests, and that the results of the this work are merely presentation of analysis of the publicly available data.

%TC:endignore  

%It should be noted that the accuracy results for majority of the methods was around 70\%. It could be concluded that there is an acceptable degree of agreeability between features and the death ratio. However, having features that have higher correlations with the output could lead to a better prediction. Further, it could be seen that regardless of the method used, the accuracy results for both training and test sets were clustered around 70\% meaning that the system is not overfitted. The only reason that prevents the accuracy to increase further arises from the fact that we have limited available data. Therefore, the future work for the current study would be to gather more data to include other states as well. Moreover, the regression version of the recommended algorithms will be used to predict the numerical value of the change in the death ratio given the selected features.

% %%%%%%%%%%%%%%%%%%%
\clearpage
\bibliographystyle{ieeetr}
\bibliography{references.bib, Ref_Miad.bib, Ref_Methods.bib}

\section{Additional Information}
There are no funding resources to be acknowledged and no conflicts of interest to disclose.
\section{Author Contributions}
A.L., M.B., S.Z, and VRH designed the project.  A.L., M.B., and S.Z collected the data, and wrote the original version of the manuscript. A.L., M.B., S.Z., and N.M. wrote the classifiers and implemented
the machine learning code. All authors contributed to the writing of the manuscript and discussion of results. VRH supervised the project. 

\end{document}